\documentclass[letterpaper, 10 pt, conference]{ieeeconf} 
\IEEEoverridecommandlockouts    
\usepackage{graphicx} 
\usepackage{booktabs} 
\usepackage{eurosym} 
\usepackage{mathptmx} 
\usepackage{times} 
\usepackage{amsmath} 
\usepackage{amssymb}  

\title{\LARGE \bf
Beyond Human-Derived Motion: Exploring Robot-Specific Execution for Construction Humanoids}

\author{MinJun Chang$^{1}$, Taehyung Kim$^{2}$, Yong K. Cho$^{3}$, and Francis Baek$^\dagger$ 
\thanks{$^{1}$MinJun Chang is with the School of Civil and Environmental Engineering, Georgia Institute of Technology, Atlanta, GA 30332, USA
        {\tt\small minjunchang@gatech.edu}}%
\thanks{$^{2}$Taehyung Kim is with the Department of Mobility Systems Engineering, Yonsei University, Seoul, South Korea
        {\tt\small kth8606@yonsei.ac.kr}}%
\thanks{$^3$Dr. Yong K. Cho is a professor at the School of Civil and Environmental Engineering, Georgia Institute of Technology, Atlanta, GA 30332, USA
        {\tt\small yong.cho@ce.gatech.edu}}%
\thanks{$^\dagger$Dr. Francis Baek is an assistant professor at the School of Civil and Environmental Engineering, Georgia Institute of Technology, Atlanta, GA 30332, USA
        {\tt\small francis.baek@ce.gatech.edu}}%
        }

\begin{document}

\maketitle
\thispagestyle{empty}
\pagestyle{empty}

\begin{abstract}

Construction productivity has remained a persistent challenge, motivating growing interest in humanoid robots that can operate in workspaces and with tools designed for human workers. Realizing this potential requires effective motion control, for which human-derived motion transfer has emerged as a promising approach by capturing and retargeting human demonstrations onto humanoid robots. Human-derived motion transfer, however, may not fully define how a humanoid should ultimately execute the transferred motion. Once task-relevant motion information has been conveyed, the humanoid may also exploit its own embodiment through motions that depart from the human demonstration. This study examines whether such Robot-Specific Motion offers an advantage in motion execution efficiency over Human-Derived Motion. On an adult-sized Unitree H1-2, Human-Derived Motion was generated through vision-based capture and retargeting, whereas Robot-Specific Motion used inverse kinematics and minimum-jerk transfers to reach the same targets while departing from the demonstrated posture, as verified using Rapid Upper Limb Assessment (RULA). Across a pick-and-stack sequence representing construction material handling, with four matched motion targets and three trials per condition, Robot-Specific Motion achieved 12.87\,s against 25.03\,s in execution time, 0.92\,m against 4.33\,m in trajectory length, and 58.7\,J against 247.3\,J in two-arm mechanical work. These findings suggest that allowing humanoids to depart from demonstrated human motion may offer additional opportunities for more efficient motion execution, while broader construction tasks are needed to determine whether such differences translate into productivity gains.
\end{abstract}

\section{Introduction}

Construction productivity has remained a persistent challenge, in part because jobsites are unstructured and task content changes daily, which limits automation that depends on fixed workcells or repeatable part presentation~\cite{bock2015future}. A further constraint is that access routes, work surfaces, and tools are dimensioned for the human body, an environment that conventional wheeled or fixed-base platforms are poorly suited to operate within. Humanoid robots have received growing attention as a potential automation approach in response, since human-like proportions allow a robot to enter such a workspace and handle tools designed around the human body without drastically redesigning the site~\cite{uthai2026humanoid, kumagai2019plasterboard}. If humanoids can participate in a range of construction operations under these conditions, they offer a route to productivity improvement that does not require restructuring the work environment itself. Translating this potential into practice requires effective approaches for generating and controlling humanoid motions to perform various construction operations.

One widely used approach for generating humanoid motion is human-derived motion transfer, in which a skilled worker's demonstration is captured and retargeted onto the robot. Two major methods carry out this transfer. In teleoperation, a human operator drives the robot in real time, sending continuous pose commands that the robot tracks as the operator performs the task~\cite{darvish2023teleoperation}. In learning-based transfer, a policy is trained offline from recorded human motion, typically video of a person performing the task, and reproduces the movement without an operator present. Such systems have extended from arm motion toward whole-body behaviors~\cite{he2024h2o, fu2024humanplus, he2024omnih2o, li2024okami}, including an application to construction tasks~\cite{construction_humanoid_2026}. These efforts establish that human-derived motion transfer can reproduce increasingly complex movement on real hardware, which makes it a practical route to acquiring the skilled techniques that experienced construction workers develop through training and site experience.

While human-derived motion transfer provides a promising way to convey task-relevant motion to humanoid robots, it may not fully define how a humanoid should ultimately execute the task. Once task-relevant motion information has been transferred, the humanoid may also use joint configurations, postures, or trajectories that differ from those demonstrated by the human. A skilled worker organizes motion around properties of the human body, keeping loads near the torso and reaching within a shoulder envelope subject to fatigue, so a demonstrated trajectory carries these characteristics alongside the task itself. Because humanoid robots differ from humans in morphology, joint structure, actuation, and reachable workspace, such alternative motions may allow the robot to better utilize its own embodiment while reaching the same motion targets. It therefore remains unclear whether preserving the human-derived motion is necessary for execution once the demonstrated motion has defined the task-relevant targets, and whether robot-specific departures can instead provide an advantage in motion execution efficiency.

The objective of this study is to examine whether Robot-Specific Motion, defined here as motion generated to reach the same motion targets using the robot's own embodiment rather than the demonstrated human movement pattern, can achieve more efficient motion execution than Human-Derived Motion. Both strategies were implemented on an adult-sized Unitree H1-2 in a laboratory environment, with scope limited to arm motion through the high-level arm interface. Robot-Specific Motion achieved shorter execution time, shorter trajectory length, and lower two-arm mechanical work while reaching the same targets, suggesting that allowing humanoids to depart from demonstrated human motion may create additional opportunities for more efficient motion execution.

\section{Methods}
\label{sec:methods}

This section describes how the two strategies are implemented on an adult-sized Unitree H1-2 and differ only in the movement pattern used to reach the motion targets. Human-Derived Motion captures, reconstructs, and retargets an observed human demonstration, while Robot-Specific Motion reaches the same targets while relaxing the postural characteristics the demonstration carries.

\subsection{Human-Derived Motion}

Human arm movement is captured with an iPhone 16 Pro running Stray Scanner~\cite{blomqvist2021stray}, recording RGB at 1920$\times$1440 and 60\,fps together with LiDAR depth at 256$\times$192 and the camera intrinsics. Three-dimensional pose is reconstructed from MediaPipe~\cite{lugaresi2019mediapipe} landmarks with confidence-gated depth. The reconstructed movement is mapped to the humanoid URDF through direction-vector matching, aligning upper-arm and forearm direction vectors in a torso frame and solving with damped Gauss-Newton optimization for shoulder pitch, roll, yaw, and elbow flexion on each arm. The resulting trajectories are streamed at 50\,Hz through the arm-overlay interface with gravity compensation while the onboard controller maintains balance, and commanded joints, executed joints, and joint torques are logged at 50\,Hz. Motion is transferred to the arms only, so locomotion and whole-body coordination fall outside the study, a restriction that applies equally to both strategies.

The postural characteristics of interest are upper-arm elevation and elbow flexion on each arm, together with the trunk and leg configuration from which the reach is performed. These are coded using Rapid Upper Limb Assessment (RULA) grand scores~\cite{mcatamney1993rula}, computed per frame and applied to the demonstration and to Human-Derived Motion. RULA maps joint angle ranges to ordinal posture categories through validated tables, weights the upper limb that this study transfers, and applies without modification to both a human demonstration and a robot trajectory, allowing them to be described on a single scale. Scoring is posture-only, with load, coupling, and activity multipliers set to zero, so the tables serve as a posture coding scheme rather than an estimate of injury risk to the robot. This first application characterizes what an alternative strategy would need to depart from. Agreement between the two series is not treated as evidence of retargeting accuracy, since distinct configurations can share a score.

\subsection{Robot-Specific Motion}

Robot-Specific Motion is designed to reach the same motion targets while relaxing the postural characteristics identified above, so that the study can test whether departing from the demonstrated configuration improves motion execution efficiency. Arm joint angles at each target are obtained by inverse kinematics on the humanoid URDF, and transfers between successive targets are generated as minimum-jerk trajectories in joint space. These transfers are executed from a lowered semi-squat stance held throughout the sequence, adopted because it places the targets within a compact arm working range, while the onboard controller maintains balance as in the other condition. Joint speeds are capped at half the maximum the platform can produce, which is below the peaks recorded in the demonstration, so that the efficiency difference could not arise from moving faster.

RULA is then applied a second time, to Robot-Specific Motion, so that the three series can be compared across the same binned segments. Beyond posture, the strategies are held to identical conditions, namely the same four motion targets in the same order, the same number of trials, and the joint-speed cap above. Trajectory length and mechanical work are invariant to time reparameterization, so they characterize the amount of motion independently of execution speed.

\section{Experimental Design and Evaluation}
\label{sec:design}

One participant with construction work experience provided the demonstration used to generate Human-Derived Motion. A single demonstrator suits this initial study because the objective is not to characterize variability across workers but to establish a controlled reference against which Robot-Specific Motion can be compared, so that any difference in execution efficiency reflects the strategies rather than variation between demonstrators. The demonstration was recorded in a laboratory environment, and the handled object was a block of fixed dimensions placed at fixed target positions.

The executed robot motion was checked against the commanded trajectory, with mean end-effector deviation of 1.2 to 2.8\,cm. Both strategies use the same 50\,Hz arm-overlay path, so this confirms that the shared execution mechanism realizes each commanded trajectory, which the efficiency comparison requires.

\begin{figure*}[!t]
\centering
\includegraphics[width=\textwidth]{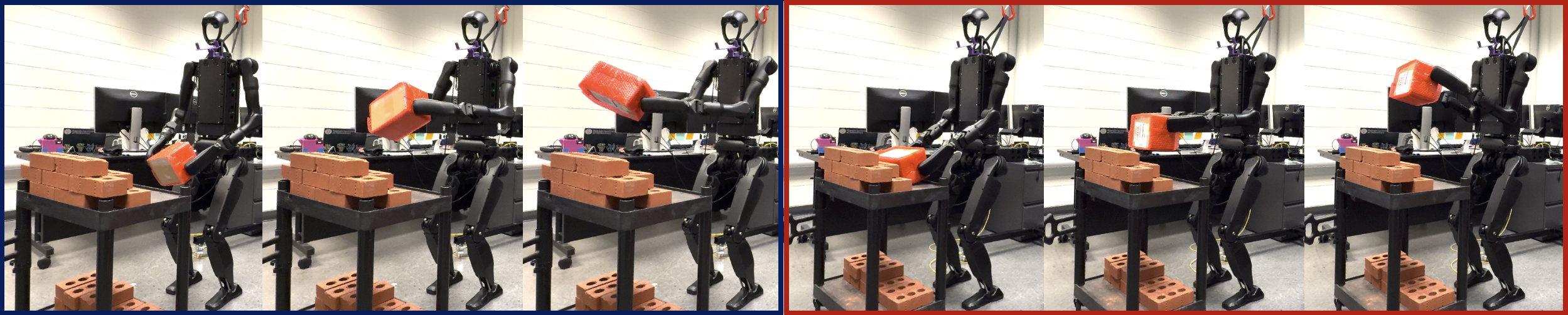}
\caption{Matched phases of the pick-and-stack sequence. Human-Derived Motion (blue) lifts high and holds its hands wider than the block, while Robot-Specific Motion (red) works from a sustained knee flexion.}
\label{fig:hardware}
\end{figure*}

The comparative experiment used a simplified pick-and-stack sequence as a controlled representation of construction material handling, which recurs across construction operations, consists of reaching, grasping, transporting, and placing materials, and requires no trade-specific judgment of the kind that would complicate a matched comparison~\cite{vaz2020humanoid, huang2023dexterous}. Matched phases of the two conditions are shown in Fig.~\ref{fig:hardware}.

\begin{figure}[t]
\centering
\includegraphics[width=\columnwidth]{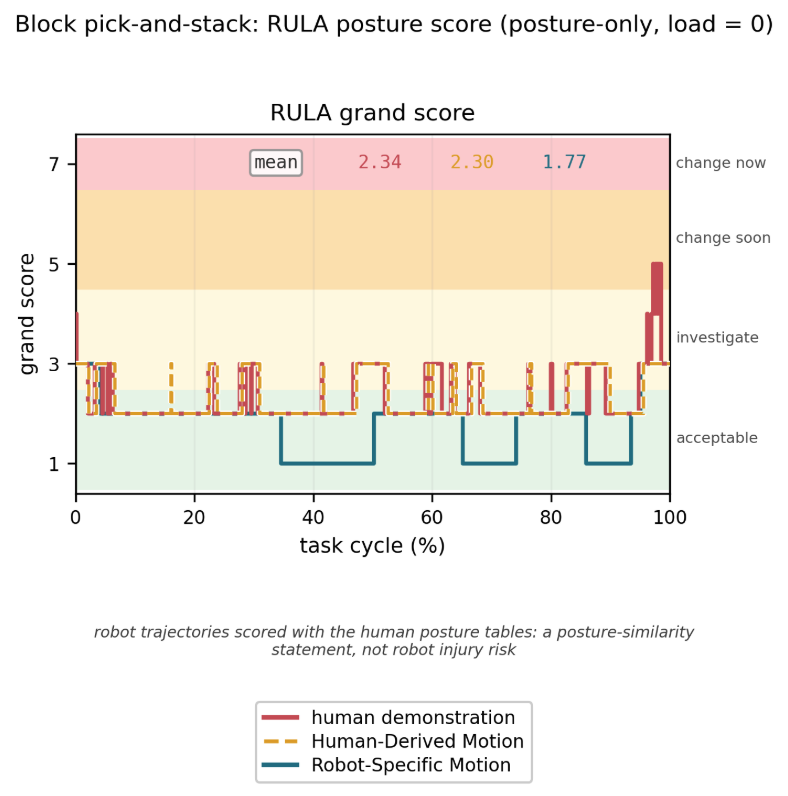}
\caption{RULA grand score over the motion sequence. Human-Derived Motion (orange) tracks the human demonstration (red), while Robot-Specific Motion (teal) holds a lower and more constant score, confirming the intended postural distinction.}
\label{fig:ergo}
\end{figure}

Three measures characterize motion execution efficiency and serve as preliminary indicators of task productivity, since characteristics of this kind could accumulate when similar motions repeat. Execution time, in seconds, characterizes temporal efficiency. Trajectory length, in meters, measured at the block center, characterizes the amount of motion required. Two-arm mechanical work, in joules, characterizes mechanical effort. RULA grand scores, dimensionless from 1 to 7, characterize departure from the demonstrated posture. The two RULA applications confirm that the strategies differed as intended. Human-Derived Motion remained close to the demonstration at a mean grand score of 2.30 against 2.34, whereas Robot-Specific Motion differed by 0.57, more than ten times as far (Fig.~\ref{fig:ergo}).

\section{Results}
\label{sec:results}

Robot-Specific Motion required less execution time, shorter trajectory length, and lower mechanical work than Human-Derived Motion under the matched target conditions, completing the sequence in 12.87\,s against 25.03\,s, over a block-center trajectory of 0.92\,m against 4.33\,m, and at 58.7\,J against 247.3\,J of two-arm mechanical work (Table~\ref{tab:pilot}). Both conditions reached all four motion targets in all three trials. Because joint speeds were capped at half the platform maximum and because trajectory length and mechanical work are invariant to time reparameterization, the differences cannot be attributed to moving faster.

\begin{figure}[t]
\centering
\includegraphics[width=0.8\columnwidth]{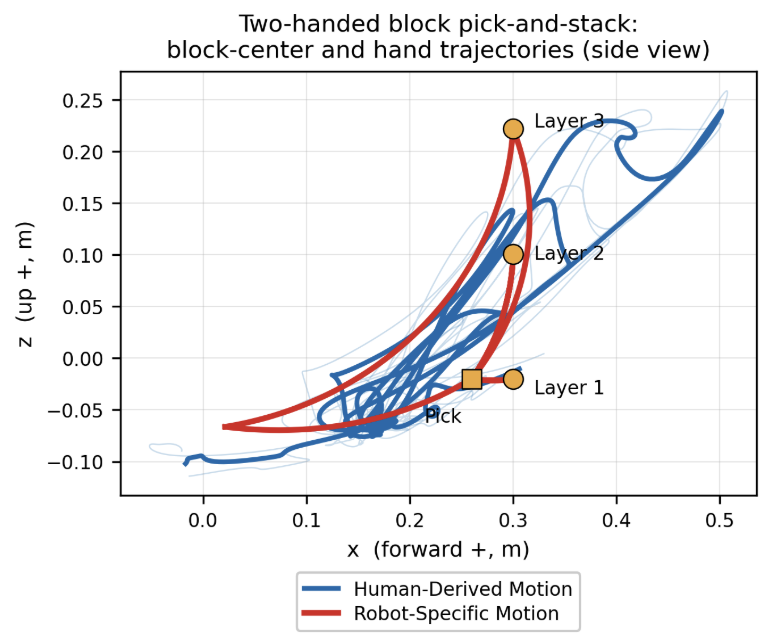}
\caption{Block-center trajectories in side view. Human-Derived Motion (blue) loops above the work zone, while Robot-Specific Motion (red) transfers directly between targets.}
\label{fig:paths}
\end{figure}

Fig.~\ref{fig:paths} shows where the difference in trajectory length originates. Human-Derived Motion re-enters the pick target eight times per trial against three for Robot-Specific Motion, so the additional path is not explained by the number of targets that had to be reached. Portions of the demonstrated movement pattern therefore persisted through retargeting without contributing to target attainment.

\begin{table}[t]
\centering
\caption{Motion execution efficiency over the pick-and-stack sequence, three trials per condition (mean $\pm$ sd)}
\label{tab:pilot}
\begin{tabular}{lccc}
\toprule
measure & Human-Derived & Robot-Specific & ratio \\
\midrule
execution time (s) & 25.03$\pm$0.06 & 12.87$\pm$0.15 & 1.9x \\
trajectory length (m) & 4.33$\pm$0.08 & 0.92$\pm$0.03 & 4.7x \\
two-arm mech.\ work (J) & 247.3$\pm$3.0 & 58.7$\pm$5.6 & $\approx$4.2x \\
\midrule
four targets reached & 3/3 & 3/3 & N/A \\
\bottomrule
\end{tabular}
\end{table}

The posture comparison indicates that this reflects a different movement strategy rather than a scaled version of the same one. Robot-Specific Motion not only received a different RULA score but also held a lower and steadier score across the task cycle, whereas Human-Derived Motion varied with the phase of the sequence in the same manner as the demonstration (Fig.~\ref{fig:ergo}). The two conditions were therefore coded into different posture categories while reaching the same four targets. Because the grand score aggregates several body segments, attributing the efficiency differences to specific postural changes would require component-level scores or joint-angle trajectories. For the evaluated motion sequence, relaxing the demonstrated postural characteristics was associated with more efficient motion execution across all three measures.

\section{Discussion and Limitations}
\label{sec:discussion}

Given the limited scope of this initial study, these findings should be interpreted as evidence of a potential design direction rather than a general conclusion about humanoid control. Within this scope, the results suggest that human-derived motion transfer and robot execution may be considered as related but distinct stages of humanoid control. Human demonstrations can provide task-relevant information, while the motion used to execute that information may be given additional freedom to reflect the robot's own embodiment. The present study does not establish which components of a demonstration should be preserved or how robot-specific motion should be generated. Rather, it motivates a broader control question, namely what information should be retained from human demonstrations and what aspects of execution can be left open to robot-specific motion generation. This distinction warrants further investigation for productivity-oriented construction robotics, where fidelity to human motion and efficient robot execution may represent different design objectives.

Several limitations bound the generality of these findings. The demonstration came from a single participant, and the comparison used a single simplified motion sequence in a laboratory environment, so the results do not establish that the same relationship holds across workers or construction operations. Three trials per condition support the reported descriptive comparison but do not permit statistical inference, and the sequence represented only a short segment of motion execution rather than a complete construction task. Mechanical work was evaluated for the arms only and therefore does not characterize whole-body energy consumption, including the energetic cost of maintaining the fixed lower-body configuration. Robot-Specific Motion was also hand-designed rather than generated through optimization, so the current comparison does not establish the range or magnitude of efficiency gains achievable through robot-specific motion generation. Future work will involve multiple participants with differing skill levels, extend the comparison to broader construction operations, and investigate motion-planning or optimization approaches for generating Robot-Specific Motion.

\section{Conclusion}
\label{sec:conclusion}

This study compared Human-Derived Motion and Robot-Specific Motion on an adult-sized humanoid to examine whether motion generated for the robot's own embodiment offers an advantage in motion execution efficiency once task-relevant motion information has been transferred from a human demonstration. Across a simplified pick-and-stack sequence representing construction material handling, Robot-Specific Motion reached the same motion targets in 12.87\,s against 25.03\,s of execution time, 0.92\,m against 4.33\,m of trajectory length, and 58.7\,J against 247.3\,J of two-arm mechanical work. These differences in execution time, trajectory length, and mechanical effort represent performance characteristics that can influence productivity when similar robotic motions are repeatedly performed in construction operations. For the evaluated sequence, the results suggest that preserving all postural characteristics carried over from the human demonstration was not necessary to reach the same motion targets efficiently. The findings motivate further investigation into humanoid control approaches that retain task-relevant information from human demonstrations while allowing robot-specific motion generation, with the potential to support more productive humanoid operation in construction.

\bibliographystyle{IEEEtran}
\bibliography{reference}

\end{document}